\documentclass[conference]{IEEEtran}

\usepackage{cite}
\usepackage{amsmath,amssymb,amsfonts, mathtools}
\usepackage{algorithmic, algorithm}
\usepackage{graphicx}
\usepackage{textcomp}
\usepackage[dvipsnames, svgnames,x11names]{xcolor}
\usepackage{hyperref}
\usepackage{comment}
\usepackage{subcaption}
\usepackage{multirow}
\usepackage{makecell} 
\usepackage{tikz} 
\usepackage{verbatim}
\usepackage{pgfplots, pgfplotstable}
\pgfplotsset{compat=1.18}
\usetikzlibrary{patterns}
\usepgfplotslibrary{groupplots, dateplot, fillbetween}

\usepackage[absolute,overlay]{textpos}

\begin{document}

\begin{textblock*}{120mm}(\dimexpr0.5\paperwidth-60mm\relax,7mm)
\centering
2025 IEEE International Conference on Agentic AI (ICA)
\end{textblock*}

\title{Coordinated Strategies in Realistic Air Combat by Hierarchical Multi-Agent Reinforcement Learning}

\author{\IEEEauthorblockN{1\textsuperscript{st} Ardian Selmonaj}
\IEEEauthorblockA{\textit{IDSIA, USI-SUPSI} \\
Lugano, Switzerland \\
ardian.selmonaj@idsia.ch}
\and
\IEEEauthorblockN{2\textsuperscript{nd} Giacomo Del Rio}
\IEEEauthorblockA{\textit{IDSIA, USI-SUPSI} \\
Lugano, Switzerland \\
giacomo.delrio@idsia.ch}
\and
\IEEEauthorblockN{3\textsuperscript{rd} Adrian Schneider}
\IEEEauthorblockA{\textit{Armasuisse S+T} \\
Thun, Switzerland \\
adrian.schneider@armasuisse.ch}
\and
\IEEEauthorblockN{4\textsuperscript{th} Alessandro Antonucci}
\IEEEauthorblockA{\textit{IDSIA, USI-SUPSI} \\
Lugano, Switzerland \\
alessandro.antonucci@idsia.ch}
}

\maketitle

\begin{abstract} Achieving mission objectives in a realistic simulation of aerial combat is highly challenging due to imperfect situational awareness and nonlinear flight dynamics. In this work, we introduce a novel 3D multi-agent air combat environment and a \emph{Hierarchical Multi-Agent Reinforcement Learning} framework to tackle these challenges. Our approach combines heterogeneous agent dynamics, curriculum learning, league-play, and a newly adapted training algorithm. To this end, the decision-making process is organized into two abstraction levels: low-level policies learn precise control maneuvers, while high-level policies issue tactical commands based on mission objectives. Empirical results show that our hierarchical approach improves both learning efficiency and combat performance in complex dogfight scenarios.
\end{abstract}


\section{Introduction}\label{sec:intro}
\emph{Reinforcement Learning}~(RL) has demonstrated remarkable potential in complex decision-making tasks. Its effectiveness is exemplified by Lockheed Martin’s work~\cite{darpa}, where an RL agent outperformed human pilots in simulated air combat scenarios, establishing it as a notable milestone in the field. Multi-agent dogfighting, a close-range aerial battle between fighter aircraft, involves rapid maneuvers, high-speed engagements, and precise coordination. Its inherent complexity arises from partial observability, nonlinear dynamics, and adversarial interactions. This makes dogfighting a highly challenging domain for \emph{Multi-Agent Reinforcement Learning}~(MARL) research, requiring precise control and high-level tactical reasoning.

To address these challenges, we develop a custom 3D multi-agent air combat environment that integrates the JSBSim flight dynamics model~\cite{jsbsim}, enabling physically accurate and aerodynamically precise simulation of aircraft behavior. To effectively learn complex maneuvers, we consider a \emph{Hierarchical} MARL~(HMARL) approach by structuring decision-making across \emph{two} levels of abstraction. Low-level policies handle continuous aircraft control, while high-level policies issue tactical commands guiding strategic course of actions and engagement decisions. This hierarchical structure supports structured and explainable reasoning, facilitating the emergence of complex aerial tactics. We further incorporate \emph{heterogeneous} agents to reflect diverse capabilities and operational roles found in real-world engagements. In summary, our main contributions are:
\begin{itemize}
    \item we present a realistic air combat environment suitable for MARL simulations under heterogeneous dynamics;
    \item we introduce an effective HMARL framework to learn advanced strategies in competitive aerial engagements;
    \item we adapt the recently introduced \emph{Simple Policy Optimization}~(SPO)~\cite{spo} algorithm to the multi-agent domain;
    \item we empirically validate our model to advocate the advantages of our design choices. Our code is available at: \href{https://github.com/IDSIA-papers/HHMARL_AirCombat}{\small{\tt{github.com/IDSIA-papers/HHMARL\_AirCombat}}}.
\end{itemize}

\section{Related Work}\label{sec:rel_work}

We focus explicitly on \emph{multi-agent} RL methods in 3D air combat environments, while the survey \cite{aircombat_survey} also includes \emph{single-agent} RL and 2D dynamics. Several existing works employ techniques that are relevant to multi-agent air combat, such as tactical reward shaping~\cite{combat_new5}, heterogeneous agents~\cite{combat_new3}, attention-based neural networks for situational awareness~\cite{combat2}, or communication mechanisms~\cite{combat_new6} to improve mission strategies. \emph{Curriculum Learning}~(CL) with gradually increasing task difficulty is applied in~\cite{combat_new8}, while enhanced coordination among agents is achieved by adapted training algorithms~\cite{combat_new10}.

The application of HMARL in defense contexts is comparatively limited. An HMARL approach that employs attention mechanisms and self-play is introduced in~\cite{hier_combat1}. Frameworks more closely related to ours appear in~\cite{hier_combat_new1, hier_combat_new2}, with the former integrating CL and the latter employing heterogeneous leader-follower agents together with JSBSim.

In this work, we introduce a complex 3D air combat environment and a training framework to learn hierarchical policies using reward shaping and cascaded league-play that gradually increases mission complexity under realistic and heterogeneous conditions. In contrast to prior efforts that are built on established RL algorithms such as \emph{Proximal Policy Optimization}~(PPO)~\cite{mappo}, we additionally adapt the recently presented SPO algorithm~\cite{spo} to the hierarchical multi-agent domain. To the best of our knowledge, this adapted setup has not yet been studied in this context and represents a significant step toward enhancing the realism of such applications.

\section{Preliminaries}\label{sec:prelim}
\subsection{Multi-Agent Reinforcement Learning}\label{sec:marl}
MARL involves $n \in \mathbb{N}$ agents learning in a shared environment. We model the agent interactions as a \emph{Partially Observable Markov Game}~(POMG)~\cite{markov_game_def}, defined by $\mathcal{G}_m$ as:
\begin{align*}
\mathcal{G}_m \coloneqq (\mathcal{N}, \mathcal{S}, \{\mathcal{A}^i\}_{i=1}^n, \{\mathcal{O}^i\}_{i=1}^n, \{O^i\}_{i=1}^n, P, \rho, \{R^i\}_{i=1}^n, \gamma)\,,
\end{align*}
where, $\mathcal{N} = \{1, \dots, n\}$ is the set of agents, $\mathcal{S}$ is the state space, $\mathcal{A}^i$ the action space of agent $i \in \mathcal{N}$, with $\boldsymbol{\mathcal{A}}=\prod_{i=1}^n \mathcal{A}^i$ being the joint action set, $\mathcal{O}^i$ is the set of possible agent observations, while $O^i: \mathcal{S} \to \Delta(\mathcal{O}^i)$ is the agent's observation function with probability simplex $\Delta(\cdot)$, $P: \mathcal{S} \times \boldsymbol{\mathcal{A}} \to \Delta(\mathcal{S})$ is the state transition kernel, $\rho \in \Delta(\mathcal{S})$ is the initial state distribution, $R^i: \mathcal{S} \times \boldsymbol{\mathcal{A}} \times \mathcal{S} \to \mathbb{R}$ is the reward function of agent $i$, and $\gamma \in [0, 1)$ controls the effect of future rewards.

At each time step $t \in \mathbb{N}$, let the random variable $S_t$ describe the state of the game. Starting from an initial state $s_0 \sim \rho$, each agent $i$ receives a local observation $o_t^i \sim O^i(\cdot \mid s_t)$ and selects an action $a_t^i \in \mathcal{A}^i$ according to its individual policy $\pi^i(a \mid o_t^i)$. The joint action $\boldsymbol{a}_t = (a_t^1, \dots, a_t^n) \in \boldsymbol{\mathcal{A}}$ is drawn from the joint policy $\boldsymbol{\pi}( \boldsymbol{a}_t \mid \boldsymbol{o}_t) = \prod_{i=1}^n \pi^i(a_t^i \mid o_t^i).$ The POMG transitions to a new state according to $s_{t+1} \sim P(s_{t+1} \mid s_t, \boldsymbol{a}_t)$, and each agent $i$ receives a reward $R^i(s_t, \boldsymbol{a}_t, s_{t+1})$. Each agent aims to maximize its expected discounted return:
\begin{equation}
    V^i(s) \coloneqq \mathbb{E}_{P, O, \boldsymbol{\pi}} \left[ \sum_{t=0}^{\infty} \gamma^t R^i(S_t, \boldsymbol{A}_t, S_{t+1} \mid S_0 = s) \right]\,.
    \label{eq:value_func}
\end{equation}


\subsection{Hierarchical Multi-Agent Reinforcement Learning}\label{sec:hrl}
HMARL uses temporal abstraction by breaking the overall task into a hierarchy of sub-tasks, which improves learning efficiency~\cite{hier_def_barto}. Abstract hierarchical commands, called an \emph{option}, invoke a control policy for a limited duration. Within a lower level, the same option can be applied to control similar aircraft, which supports generalization and alignment with hierarchical structures of defense organizations. Our hierarchical system is modeled as a \emph{Partially Observable Semi-Markov Game}~(POSMG)~\cite{semi_mg}, given as:
\begin{align*}
\mathcal{G}_s \coloneqq (\mathcal{N}, \mathcal{S}, \{\mathcal{C}^i\}_{i=1}^n, \{\mathcal{Z}^i\}_{i=1}^n, \{Z^i\}_{i=1}^n, \mathcal{P}, \rho, \{\mathcal{R}^i\}_{i=1}^n, \gamma)\,,
\end{align*}
with $\mathcal{N}, \mathcal{S}, \rho, \gamma$ as for $\mathcal{G}_m$, while the new components are: $\mathcal{C}^i$, the set of temporally extended high-level actions (options) available to agent $i$. Each option $c \in \mathcal{C}^i$ is a triple $c=(\mathcal{I}_c, \pi_{c}, \beta_c)$, with initiation set $\mathcal{I}_c \subseteq \mathcal{S}$, intra-option controller $\pi_{c}$ (being a low-level policy from $\mathcal{G}_m$), and termination condition $\beta_c: \mathcal{S} \rightarrow [0,1]$. The observation function is $Z^i: \mathcal{S} \to \Delta(\mathcal{Z}^i)$, where $\mathcal{Z}^i$ is the observation space. The semi-Markov transition kernel over next states and option duration is given by $\mathcal{P}: \mathcal{S} \times \boldsymbol{\mathcal{C}} \rightarrow \Delta(\mathcal{S} \times \mathbb{N}_{+})$, with joint option set $\boldsymbol{\mathcal{C}} = \prod_i \mathcal{C}^i$. The cumulative reward is given by $\mathcal{R}^i: \mathcal{S} \times \boldsymbol{\mathcal{C}} \times \mathcal{S} \times \mathbb{N}_{+} \rightarrow \mathbb{R}$. 
At each decision epoch $(t_k, k) \in \mathbb{N}$ with random durations $\tau_k = t_{k+1}- t_k \in \mathbb{N}_+$, agent $i$ observes $z_{t_k}^i \sim Z^i(\cdot \mid s_{t_k})$ and selects an option according to its high-level policy $c_{t_k}^i \sim \pi_h^i(\cdot \mid z_{t_k}^i)$, which activates one of the low-level policies $\{\pi_{c_{t_k}^j}\}_{j=1}^n$ to execute primitive-time actions $a_t \sim \pi_{c_{t_k}}(\cdot \mid o_t)$ in the underlying POMG until at least one option terminates according to $\{\beta_{c_{t_k}^j}\}_{j=1}^n$ after duration $\tau_k$. The state evolves as $(s_{t_{k+1}}, \tau_k) \sim \mathcal{P}(\cdot \mid s_{t_k}, \boldsymbol{c}_{t_k})$ and a cumulative reward $\mathcal{R}^i(s_{t_k}, \boldsymbol{c}_{t_k}, s_{t_{k+1}}, \tau_k)$ is distributed to each agent. The objective is to maximize the expected return~\eqref{eq:value_func}.

\section{Method}\label{sec:method}
\subsection{Environment}\label{sec:env}
\begin{figure}[htb!]
    \centering
    \begin{subfigure}[b]{0.57\columnwidth}
        \includegraphics[width=\linewidth]{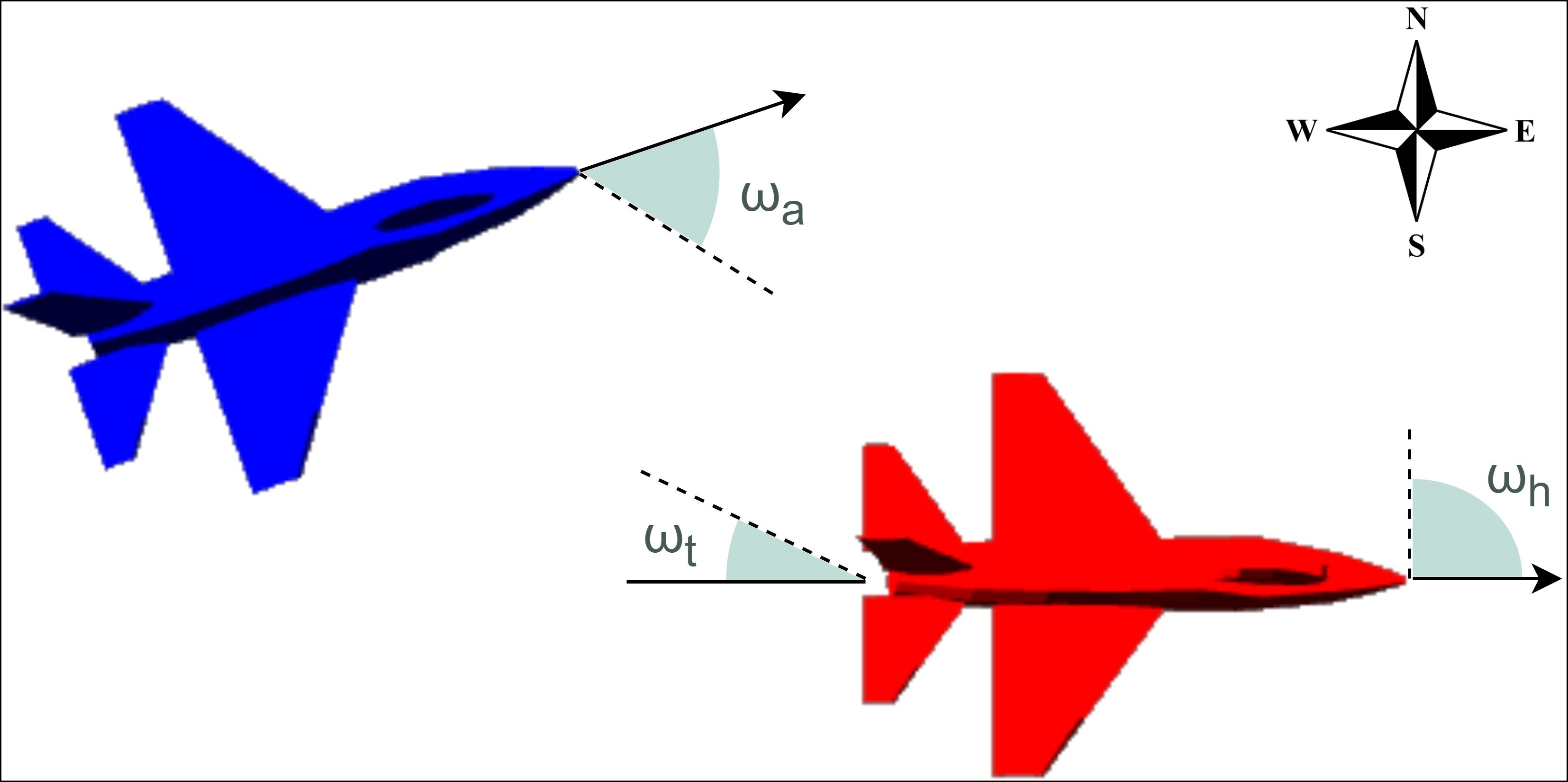}
        \caption{Orientation angles.}
        \label{fig:env_angles}
    \end{subfigure}
    \hspace{1cm}
    \begin{subfigure}[b]{0.16\columnwidth}
        \includegraphics[width=\linewidth]{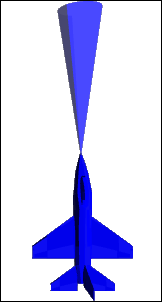}
        \caption{WEZ.}
        \label{fig:WEZ}
    \end{subfigure}
    \caption{Aircraft postures showing: (a) attacking angles, and (b) cannon fire modeled through the WEZ.}
    \label{fig:env_metrics}
\end{figure}

Air combat scenarios are generally categorized into \emph{beyond visual range} and \emph{within visual range} engagements. Here, we focus on the latter, that is the relevant scenario for dogfighting. Realistic aircraft physics is essential for competitive performance. Therefore, we built a custom training environment using JSBSim, an open-source flight dynamics simulator based on real-world physics and control systems. An effective MARL process requires careful design of agent interactions, state–action spaces, reward structures, and scalability to capture mission dynamics. In our environment, the state space includes the aircraft’s geographic position $(x, y, z) \, [\text{m}]$, velocity $v \, [\text{m/s}]$, acceleration $\dot{v} \, [\text{m/$\text{s}^2$}]$, relative distance to opponents $d \, [\text{m}]$, \emph{Aspect Angle}~(AA) $\omega_a \, [^\circ]$, \emph{Antenna Train Angle}~(ATA) $\omega_t \, [^\circ]$, and heading angle $\omega_h \, [^\circ]$ (Fig.~\ref{fig:env_angles}). All scalar values of state variables are normalized to the range $[-1, 1]$ and concatenated to a vector. The action space is different for each hierarchy level and presented in the following. Cannon fire, i.e., an unguided kinetic weapon, is modeled using a \emph{Weapon Engagement Zone}~(WEZ), in which an agent has an $80\%$ probability of being destroyed per simulation step. The WEZ covers a range of $3.5 \,\mathrm{km}$ and an angular span of $8^\circ$ (Fig.~\ref{fig:WEZ}). Missiles are excluded, as their long range and guidance capabilities could negate the purpose of close-range engagements for dogfighting.

As real-world combat scenarios often involve diverse aircraft types, we consider an heterogenenous setup involving the established F16 Fighting Falcon and Douglas A4 Skyhawk models. The F16 is a supersonic, highly maneuverable fighter, whereas the A4 Skyhawk is a subsonic aircraft with lower top speed but excellent handling at low altitudes. We keep the WEZ constant for both aircraft, while their differing flight dynamics are realistically simulated through JSBSim, which runs at a physics integration frequency of $100$ Hz, i.e., $\delta t = 0.01 \, \mathrm{seconds}$ are simulated per environment step. 

\subsection{Low-Level Control Policies}\label{sec:low-level-policies}
Low-level policies are responsible for steering the aircraft, of which we define three types in the following. The actions $a^i_{t} \in \mathcal{A}^i = [0,1]^4$ of each agent $i$ include the maneuver controls: \emph{i)}~aileron (roll), \emph{ii)}~elevator (pitch), \emph{iii)}~rudder (yaw), \emph{iv)}~throttle for power management, and \emph{v)}~a shoot command for cannon fire. Their observation $o^i_t$ contains information about themselves and their nearest enemy aircraft. They constitute the intra-option controllers $\pi_c$ of $\mathcal{G}_s$ to be selected by the high-level policy $\pi_h$. Based on domain knowledge, we restrict our framework to a small set of low-level policies, which revealed to be a reasonable design choice as they reflect core tactical maneuvers in real-world engagements, where simplicity and robustness are favored over an overly diverse action set. Similar low-level polices are also considered by~\cite{darpa}~and~\cite{icmla}.

\subsubsection{Engage} 
The engage policy $\pi_e$ aims to achieve an advantageous position by placing the agent behind the enemy, pointing towards it while the enemy is oriented straight ahead. Let $\omega_r$ denote the roll angle. We model deviations from optimal values $\omega_a = \omega_t = \omega_r = 0$ as Gaussian functions:
\begin{equation}
    b_a \coloneqq e^{-(\frac{\omega_a}{10 ^\circ})^2}, \quad b_t \coloneqq e^{-(\frac{\omega_t}{10^\circ})^4}, \quad b_r \coloneqq e^{-(\frac{\omega_r}{25^\circ})^2}\,.
    \label{eq:heading_rew_1}
\end{equation}
Function $b_t$ has a steeper slope to emphasize the importance of positioning behind the enemy. Similarly, the distance component $b_d$ decreases if the agent is outside shooting range:
\begin{equation}
    b_d \coloneqq
    \begin{cases}
    e^{-(\frac{d- 3.5 \, \mathrm{km}}{10 \, \mathrm{km}})^2}, & \text{if } d > 3.5 \, \mathrm{km}\,,\\
    1, & \text{if } d \leq 3.5 \, \mathrm{km}\,.
    \end{cases}
    \label{eq:heading_rew_2}
\end{equation}
The total posture reward is defined as the smoothed product:
\begin{equation}
    r_p \coloneqq \sqrt{b_a \cdot b_t \cdot b_r \cdot b_d} \quad \in [0,1]\,.
    \label{eq:heading_rew_engage}
\end{equation}
With an additional event reward $r_e \coloneqq -5$ for crashing (i.e., hitting the ground or leaving map boundaries) or being destroyed by the opponent, and a shooting penalty $r_s\coloneqq-0.1$ per shot, the total per-step reward for $\pi_e$ is defined as:
\begin{equation}
    r_{\pi_e} \coloneqq r_s + r_e + 0.05 \cdot r_p \quad \in [-5.1, 0.05]\,.
\end{equation}

\subsubsection{Attack} 
The attack policy $\pi_a$ aims to aggressively destroy the enemy. We model the kill reward based on the positioning at the moment the enemy is destroyed, i.e., the more the agent attacks from behind, the higher the reward:
\begin{equation}
    r_k \coloneqq (g_{1,10} \circ \cos)(\omega_t) = 4.5\cos(\omega_t) + 5.5 \quad \in [1, 10]\,,
    \label{eq:kill_rew}
\end{equation}
where $g_{\mathrm{a,b}}(\cdot)$ is a range transformation function that shifts the original range to $[\mathrm{a}, \mathrm{b}]$. We also include a slightly relaxed posture reward to encourage the agent to approach the enemy:
\begin{equation}
    r_o \coloneqq \sqrt{b_a \cdot b_r \cdot b_d} \quad \in [0,1]\,.
    \label{eq:orientation_rew_engage} 
\end{equation}
The final per-step reward used to train $\pi_a$ is given by:
\begin{equation}
    r_{\pi_a} \coloneqq r_s + r_e + r_k + 0.05 \cdot r_o \quad \in [-5.1, 10.05]\,.
\end{equation}

\subsubsection{Defend} 
The defend policy $\pi_d$ aims to evade opponents by maintaining a large distance from them. To achieve effective evasive behavior, we use a potential reward:
\begin{equation}
    r_d \coloneqq \Phi(s_{t+1}) - \Phi(s_t), \quad \Phi(s_t) \coloneqq \frac{d(s_t)}{d_m}\,,
    \label{eq:pot_reward}
\end{equation}
where $d(s_t)$ is the distance to the enemy at state $s_t$, and $\Phi(s_t)$ is normalized by the map size $d_m$. Therefore, $r_d$ is positive only if the distance at the new state $s_{t+1}$ is larger than at the previous state $s_t$. The total per-step reward for $\pi_d$ is:
\begin{equation}
    r_{\pi_d} \coloneqq r_s + r_e + r_d \quad \in [-5.1, 0.1]\,.
\end{equation}

\subsection{High-Level Commander Policy}\label{sec:high-level-policy}
The \emph{commander} policy $\pi_h$ determines which low-level policy each agent will use. The observations $z^i_{t_k}$ include information about \emph{i)}~the calling agent $i$, \emph{ii)}~its two closest opponents, and \emph{iii)}~its nearest friendly aircraft. The option set is $\mathcal{C}^i \coloneqq \{0, 1, 2\}, i \in \mathcal{N}$, where $c_{t_k}=0$ activates $\pi_d$, $c_{t_k}=1$ activates $\pi_e$, and $c_{t_k}=2$ activates $\pi_a$. The selected low-level policy $\pi_c$ then receives the corresponding observation $o^i_t$. While the commander has no direct control over the agent's steering maneuvers, it is responsible for deciding an effective course of actions for its low-level policies. We keep the reward structure simple, allowing $\pi_h$ to autonomously discover an optimal strategy. The only components are the kill reward~\eqref{eq:kill_rew} (with range $g_{1,5}$) and the event penalty $r_e = -5$, accumulated over option duration $\tau_k$. Thus, the total reward for $\pi_h$ is:
\begin{equation}
    r_{\pi_h} \coloneqq r_e + r_k \quad \in [-5, 5]\,.
\end{equation}

\begin{figure}[htb!]
    \centering
    \includegraphics[width=\linewidth]{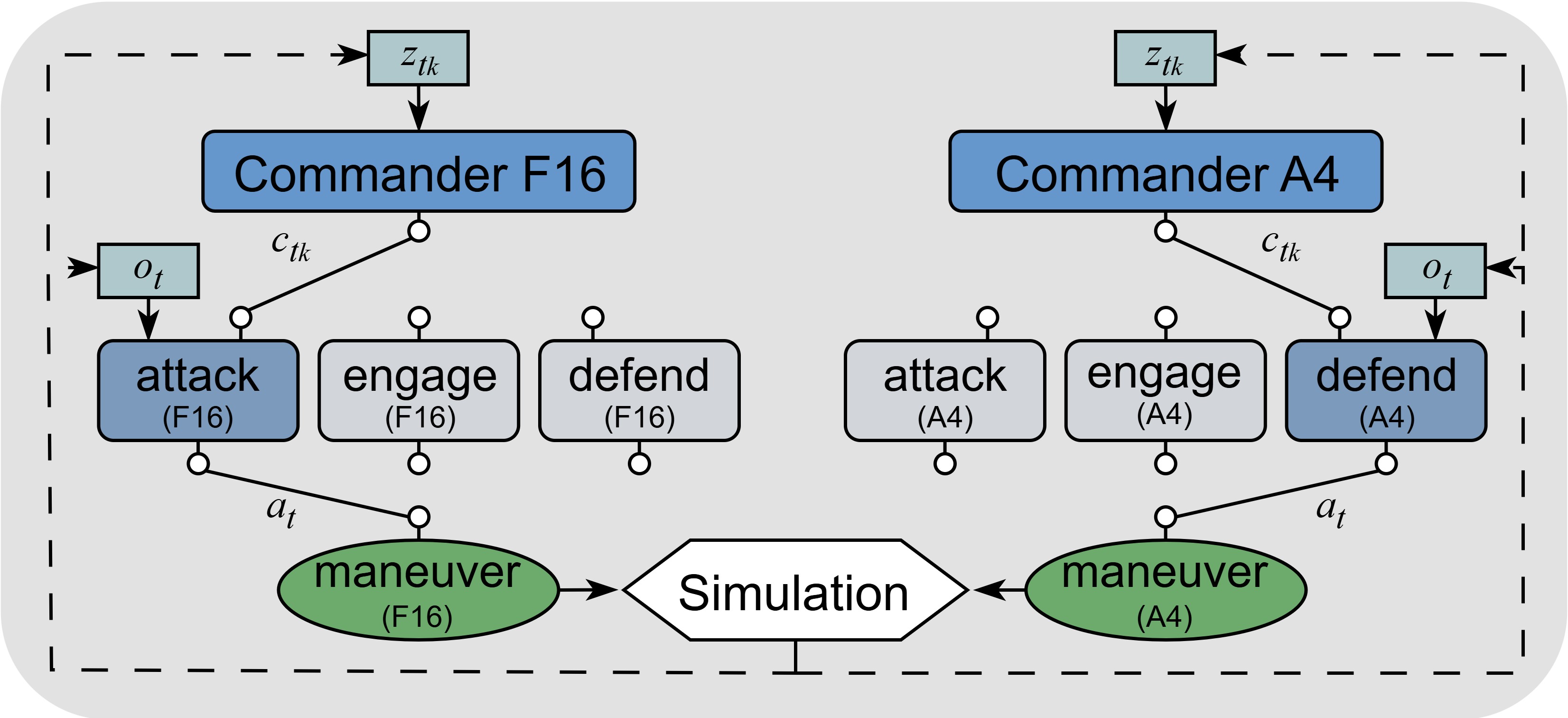}
    \caption{Hierarchical simulation process: for each aircraft type, a shared commander policy decides which low-level policy to activate, this in turn producing control maneuvers.}
    \label{fig:sim_process}
\end{figure}

\subsection{Curriculum Learning and League-Play}\label{sec:curriculum}
In our framework, we employ shared policies for each aircraft type, i.e., all F16 agents use the same instances of the low-level policies $\pi_a$, $\pi_e$, and $\pi_d$, as well as the same commander instance $\pi_h$ (Fig.~\ref{fig:sim_process}). The same also applies to A4 agents. For convenience, we abbreviate F16 with the superscript $\text{(f)}$ and A4 with $\text{(a)}$. Consequently, there are six low-level policies to learn:
$\pi_{c} \in \{\pi^{\text{f}}_{a}, \pi^{\text{f}}_{e}, \pi^{\text{f}}_{d}, \pi^{\text{a}}_{a}, \pi^{\text{a}}_{e}, \pi^{\text{a}}_{d} \},$
and two high-level policies:
$\pi_h \in \{\pi^{\text{f}}_{h}, \pi^{\text{a}}_{h} \}.$

The learning process starts with training low-level policies. We adopt a CL scheme with levels by transferring the policy to the next level once training completes. Level complexity increases by making opponents more competitive. We first employ a baseline aircraft that follows random pre-set locations, similar to $\pi_e$ but with reduced observation and action spaces, denoted $\pi_{\text{0}}$. The enemy behavior according to CL levels are: \emph{L1)} $\pi_{\text{0}}$ with random targets, \emph{L2)} $\pi_{\text{0}}$ targeting the agent with slight disturbances and manual cannon fire, \emph{L3)}~self-play, where the agent engages against itself, and \emph{L4)}~league-play, where each control policy $\pi_c$ is trained against mixed opponent behaviors, i.e., in each episode, opponents are randomly assigned any $\pi_c$ from L3. This final stage L4 further improves engagement capabilities and prepares agents for diverse opponent strategies. Once low-level training completes, we train high-level policies $\pi_h$ with $\pi_c$ fixed as options. CL is no longer required, since low-level skills are already established and the focus shifts to coordinating them strategically. Instead, $\pi_h$ faces opponents with a mixed strategy over the L4 policies: $\pi_a$ ($40\text{\footnotesize{\%}}$), \, $\pi_e$ ($40\text{\footnotesize{\%}}$), and $\pi_d$ ($20\text{\footnotesize{\%}}$), thereby encouraging competitive adversary behavior.

\subsection{Learning Algorithm}\label{sec:learning_algo}
We adopt Actor-Critic neural networks~\cite{actor_critic}, where the actor models the policy $\pi$ with parameters $\theta$ and the critic, with parameters $\phi$, learns the value function defined in~\eqref{eq:value_func}. The network architecture is identical across all policies (except for observation and action spaces) and designed as follows: \emph{(i)} two linear layers of size 200 with $\text{tanh}$ activation; \emph{(ii)} an attention module (for situational awareness) of size 200; \emph{(iii)} the outputs of the three preceding components are added and passed through a normalization layer; and \emph{(iv)} a final linear output layer of size 200. The hierarchical policies $\pi_h$ follow the \emph{Centralized Training and Decentralized Execution}~(CTDE)~\cite{ctde} paradigm, where the critic has access to the full state $s_{t_k}$, while the actor operates solely on local observations $z_{t_k}$.

We employ the SPO algorithm to update network parameters. The SPO surrogate policy loss maintains policy updates within a trust region by penalizing deviations of the probability ratio. Unlike PPO, which can set gradients to zero for some samples by ratio clipping, the SPO loss uses all samples to update the policy toward the trust region. It is defined as:
\begin{equation}
    \mathcal{L}_{s} \coloneqq -\mathbb{E}_{\pi_{\theta_{old}}} \left[ r_t(\theta) \, \hat{A}_t \, - \, \frac{\lvert \hat{A}_t \rvert}{2 \epsilon} \cdot \left(r_t(\theta)-1 \right)^2 \right]\,,
    \label{eq:loss_spo}
\end{equation}
where $r_t(\theta) = \frac{\pi_\theta(a_t|s_t)}{\pi_{\theta_{\text{old}}}(a_t|s_t)}$ denotes the probability ratio, $\epsilon$ is a hyperparameter, and $\hat{A}_t$ the estimated advantage~\cite{adv_est}.
The value function loss measures the error between the predicted values $V_{\phi}(s_t)$ and the target returns $\hat{R}_t$ as follows:
\begin{equation}
   \mathcal{L}_v \coloneqq \mathbb{E}_t \left[ \left( V_\phi(s_t) - \hat{R}_t \right)^2 \right]\,. 
   \label{eq:loss_value}
\end{equation}
Finally, an entropy term encourages exploration as follows:
\begin{equation}
    \mathcal{L}_e \coloneqq -\mathbb{E}_{\pi_{\theta}} \left[ -\log \pi_\theta(a_t|s_t) \right] = -\mathcal{H}(\pi_{\theta})\,.
    \label{eq:loss_entropy}
\end{equation}
The total actor-critic loss is thus defined as:
\begin{equation}
    \mathcal{L} \coloneqq \mathcal{L}_{s} + \kappa_v \, \mathcal{L}_v + \kappa_e \, \mathcal{L}_e\,,
    \label{eq:loss_total}
\end{equation}
where coefficients $\kappa_v$ and $\kappa_e$ control the regularization terms. The pseudo-code in Algorithm~\ref{algo:spo} is used to train all policies $\pi_c$ and $\pi_h$. As the hierarchical policies $\pi_h$ employ CTDE, the actors (parameters $\theta$) independently optimize the SPO loss~\eqref{eq:loss_total} from local trajectories, and centralized critics (parameters $\phi$) estimate the value~\eqref{eq:value_func} from the global state. This constitutes the first adaptation of SPO to the multi-agent setting, denoted as \emph{Multi-Agent Simple Policy Optimization}~(MA-SPO). Motivated by the well-established adaptation of PPO to MA-PPO~\cite{mappo} and its demonstrated effectiveness in various multi-agent domains, MA-SPO is likewise expected to achieve strong performance, which we validate in the next section.

\begin{algorithm}[htb!]
\caption{Simple Policy Optimization Algorithm}
\begin{algorithmic}[1]
\STATE \textbf{Initialize:} policy $\pi_\theta$ and value $V_\phi$, hyperparameter $\epsilon$, coefficients $\kappa_v, \kappa_e$, learning rates $\lambda_{\theta}, \lambda_{\phi}$
\WHILE{not termination condition}
    \STATE Collect data $\mathcal{D} = \{(s_t, a_t, r_t)\}_{t=1}^M$ using policy $\pi_\theta$
    \STATE Set $\pi_{\theta_{\text{old}}} \leftarrow \pi_\theta \quad \text{and} \quad V_{\phi_{\text{old}}} \leftarrow V_\phi$
    \STATE Estimate advantage $\hat{A}(s_t, a_t)$ through~\cite{adv_est}
    \STATE Compute return $\hat{R}_t \leftarrow V_{\phi_{\text{old}}}(s_t) + \hat{A}(s_t, a_t)$
    \FOR{each training epoch}
        \STATE Compute $\mathcal{L}$ as by~\eqref{eq:loss_total}
        \STATE Update $\; \theta \leftarrow \theta - \lambda_\theta \nabla_\theta \mathcal{L}, \quad \phi \leftarrow \phi - \lambda_\phi \nabla_\phi \mathcal{L}$
    \ENDFOR
\ENDWHILE
\end{algorithmic}
\label{algo:spo}
\end{algorithm}

\section{Experiments}\label{sec:exp}
We conduct experiments to validate our method. In each episode, teams are randomly assigned to a side of the map with random initial postures, ensuring at least one of each aircraft type (F16 and A4) per team. Low-level agents follow the dynamics of $\mathcal{G}_m$ and are trained in a $1\text{-vs-}1$ setup, as observing additional entities may introduce noise and hinder precise maneuver control. High-level commanders are trained under $\mathcal{G}_s$ dynamics in $3\text{-vs-}3$ scenarios, with observations defined in Sec.~\ref{sec:high-level-policy}. The termination condition $\beta_c$ is either the destruction of an agent (by cannon or boundary exit) or expiration of the option duration, which is set to an upper bound of $\tau_k=15$ low-level executions.

\begin{figure}[htb!]
    \centering
    \begin{subfigure}[b]{0.485\columnwidth}
        \centering
        \includegraphics[width=\linewidth]{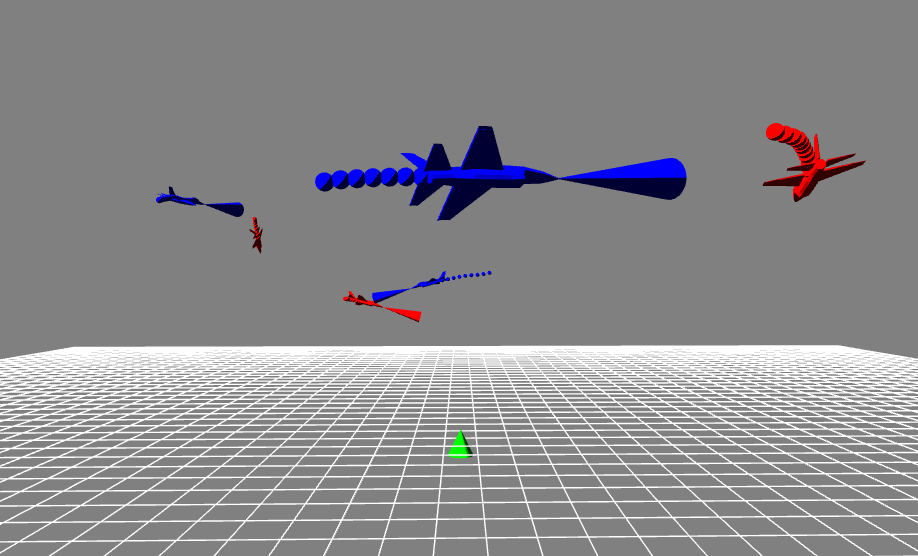}
        \caption{PyGame.}
        \label{fig:pygame}
    \end{subfigure}
    \begin{subfigure}[b]{0.485\columnwidth}
        \centering
        \includegraphics[width=\linewidth]{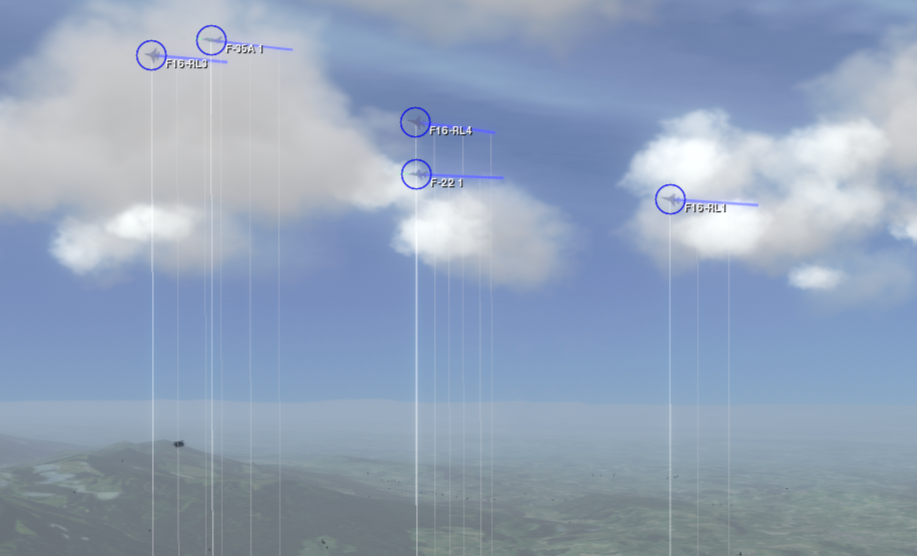}
        \caption{VR-Forces.}
        \label{fig:vrf}
    \end{subfigure}
    \caption{Rendered air combat trajectories in (a) our custom environment using PyGame, and (b) VR-Forces when deployed.}
    \label{fig:env_rendering}
\end{figure}

A rendered $3\text{-vs-}3$ air combat situation of our environment is shown in Fig.~\ref{fig:pygame}, with blue aircraft as the learning agents and red as opponents. The modularity of our environment allows to run any $n\text{-vs-}n$ combat configuration. Fig.~\ref{fig:vrf} illustrates trained agents during deployment on \emph{VR-Forces}\footnote{\href{https://www.mak.com/mak-one/apps/vr-forces}{\tt{mak.com/vr-forces}}.}~(VR-F), a popular defense simulation tool. The realism achieved by JSBsim is further enhanced by VR-F through diverse scenarios and entities. After training agents in our JSBSim environment, their deployment is established through the \emph{Distributed Interactive Simulation}~\cite{disoriginal}, enabling real-time synchronization between JSBSim and VR-F over UDP/IP sockets. For the experiments discussed in this section, however, we only used our custom environment, which is much faster than VR-F. The complete integration with VR-F is ongoing and left as a future work.

\subsection{Learning and Evaluation Parameters}
Our training uses \emph{Ray RLlib}. The SPO parameters are set as: learning rates $\lambda_{\theta} = \lambda_{\phi} = 10^{-4} \xrightarrow{15\text{M}} 10^{-5}$ (i.e., linearly decayed over $15 \text{M}$ samples and then held constant), discount $\gamma = 0.995$, ratio $\epsilon=0.25$, value coefficient $\kappa_v = 0.9$, entropy coefficient $\kappa_e = 0.05 \xrightarrow{3\text{M}} 0$, batch = $6'000$ for $\pi_c$ and $3'000$ for $\pi_h$, with mini-batches of $256$ each, and $\text{Adam}$ as optimizer. 

As SPO is similar to PPO, we compare these two \emph{on-policy} algorithms using identical parameters. In addition, we include the \emph{off-policy} algorithm \emph{Soft Actor Critic}~(SAC)~\cite{sac}, which is also used for aircraft control. Unlike SPO and PPO, which use entropy as a regularizer, SAC optimizes a maximum-entropy objective to encourage exploration using past data, making it more sample-efficient but also more sensitive to hyperparameter tuning, which can mislead policy optimization in complex control domains, as proven by~\cite{MaxEntMislead}. We keep network architectures identical across all algorithms (see Sec.~\ref{sec:learning_algo}) and slightly adapt the standard SAC parameters from \emph{Ray} to achieve best performance. Key changes include constant learning rates of $\lambda_{\theta} = \lambda_{\phi} = 3 \cdot 10^{-5}$, automatic target entropy tuning (matching the $\kappa_e$ schedule), and soft update coefficient of $0.008$, as SAC employs two critic networks to mitigate overestimation bias. Other algorithms, such as the popular reward-decomposition method QMIX~\cite{qmix}, are omitted since our approach assigns per-agent rewards. Training on a $32$-core $3.6 \, \mathrm{GHz}$ CPU required around $10 \, \mathrm{h}$ per initial $\pi_c$ level, rising up to $30 \, \mathrm{h}$ per remaining level and per commander $\pi_h$.

Evaluations are done for $1'000$ episodes. A \emph{win} is when all opponents are destroyed, \emph{loss} if all agents are dead, and \emph{draw} if none of the two cases occurs. Figures~\ref{fig:low-level-training-results} and~\ref{fig:hier_training} show the mean rewards on y-axis over training samples on x-axis.

\subsection{Low-Level Results}\label{sec:low-level-results}
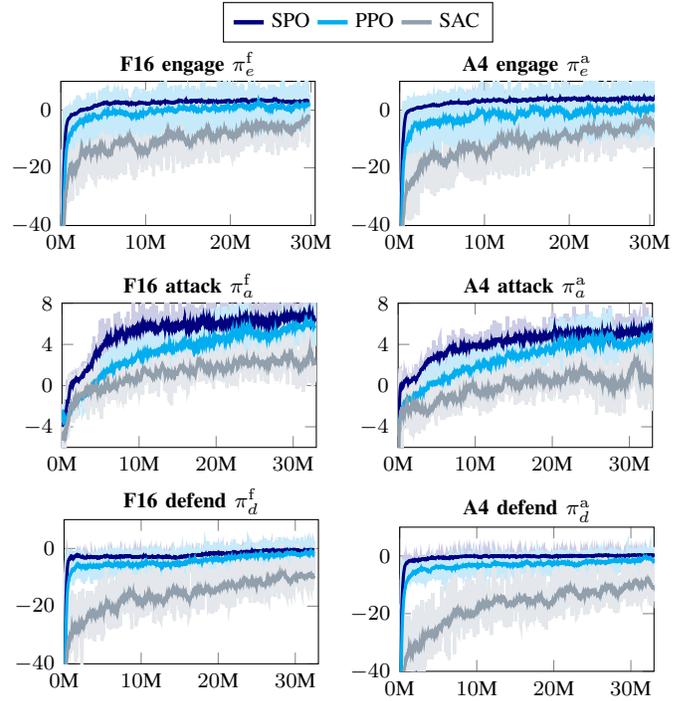
\begin{figure}[htb!]
\centering

    \begin{subfigure}{0.47\columnwidth}
    \centering
    \hspace*{-0.3cm}
    \begin{tikzpicture}
      \begin{axis}[
            title={F16 engage $\pi^{\mathrm{f}}_{e}$},
            title style={font=\bfseries\footnotesize, yshift=-0.75em},
            width=\columnwidth+0.8cm, height=3.5cm,
            xmin=-1, xmax=30000000,
            ymin=-40, ymax=10,
            xtick={0,10000000,20000000,29500000},
            xticklabels={$0\text{M}$, $10\text{M}$, $20\text{M}$, $30\text{M}$},
            scaled x ticks=false,
            ytick distance=20,
            enlargelimits=false,
            ticklabel style={font=\footnotesize},
            every axis plot/.append style={thick},
            legend style={
                at={(1.7,1.55)},
                anchor=north east,
                draw=black, 
                fill=white,  
                font=\footnotesize,
            },
            legend columns=3,
            legend image post style={line width=1.7pt}, 
            legend image code/.code={
                \draw[#1] (0cm,0cm) -- (0.4cm,0.0cm); 
            }
        ]

        \addplot[
          LightSlateGray!20,
          very thick,
          forget plot,
          name path=rew_sac
        ] table [x=samples, y=rew_sac] {exp/L1_track_F16.dat};

        \addplot[
          NavyBlue!20,
          very thick,
          forget plot,
          name path=rew_spo
        ] table [x=samples, y=rew_spo] {exp/L1_track_F16.dat};

        \addplot[
          cyan!20,
          very thick,
          forget plot,
          name path=rew_ppo
        ] table [x=samples, y=rew_ppo] {exp/L1_track_F16.dat};

        \addplot[
          NavyBlue,
          very thick,
          name path=smoothed_spo
        ] table [x=samples, y=smoothed_spo] {exp/L1_track_F16.dat};
        \addlegendentry{SPO};

        \addplot[
          cyan,
          very thick,
          name path=smoothed_ppo
        ] table [x=samples, y=smoothed_ppo] {exp/L1_track_F16.dat};
        \addlegendentry{PPO};

        \addplot[
          LightSlateGray!80,
          very thick,
          name path=smoothed_sac
        ] table [x=samples, y=smoothed_sac] {exp/L1_track_F16.dat};
        \addlegendentry{SAC};
      \end{axis}
    \end{tikzpicture}
    \end{subfigure}
    \begin{subfigure}{0.47\columnwidth}
    \centering
    \begin{tikzpicture}
      \begin{axis}[
            title={A4 engage $\pi^{\mathrm{a}}_{e}$},
            title style={font=\bfseries\footnotesize, yshift=-0.75em},
            width=\columnwidth+0.8cm, height=3.5cm,
            xmin=-1, xmax=30000000,
            ymin=-40, ymax=10,
            xtick={0,10000000,20000000,29500000},
            xticklabels={$0\text{M}$, $10\text{M}$, $20\text{M}$, $30\text{M}$},
            scaled x ticks=false,
            ytick distance=20,
            enlargelimits=false,
            ticklabel style={font=\footnotesize},
            every axis plot/.append style={thick},
        ]

        \addplot[
          LightSlateGray!20,
          very thick,
          forget plot,
          name path=rew_sac
        ] table [x=samples, y=rew_sac] {exp/L1_track_A4.dat};

        \addplot[
          NavyBlue!20,
          very thick,
          forget plot,
          name path=rew_spo
        ] table [x=samples, y=rew_spo] {exp/L1_track_A4.dat};

        \addplot[
          cyan!20,
          very thick,
          forget plot,
          name path=rew_ppo
        ] table [x=samples, y=rew_ppo] {exp/L1_track_A4.dat};

        \addplot[
          NavyBlue,
          very thick,
          name path=smoothed_spo
        ] table [x=samples, y=smoothed_spo] {exp/L1_track_A4.dat};

        \addplot[
          cyan,
          very thick,
          name path=smoothed_ppo
        ] table [x=samples, y=smoothed_ppo] {exp/L1_track_A4.dat};

        \addplot[
          LightSlateGray!80,
          very thick,
          name path=smoothed_sac
        ] table [x=samples, y=smoothed_sac] {exp/L1_track_A4.dat};
        
      \end{axis}
    \end{tikzpicture}
    \end{subfigure}

    \begin{subfigure}{0.47\columnwidth}
    \centering
    \vspace{0.1cm}
    \begin{tikzpicture}
      \begin{axis}[
            title={F16 attack $\pi^{\mathrm{f}}_{a}$},
            title style={font=\bfseries\footnotesize, yshift=-0.75em},
            width=\columnwidth+0.8cm, height=3.5cm,
            xmin=-1, xmax=33000000,
            ymin=-6, ymax=8,
            xtick={0,10000000,20000000,30000000},
            xticklabels={$0\text{M}$, $10\text{M}$, $20\text{M}$, $30\text{M}$},
            scaled x ticks=false,
            ytick distance=4,
            enlargelimits=false,
            ticklabel style={font=\footnotesize},
            every axis plot/.append style={thick},
        ]

        \addplot[
          NavyBlue!20,
          very thick,
          forget plot,
          name path=rew_spo
        ] table [x=samples, y=rew_spo] {exp/L1_cannon_F16.dat};

        \addplot[
          cyan!20,
          very thick,
          forget plot,
          name path=rew_ppo
        ] table [x=samples, y=rew_ppo] {exp/L1_cannon_F16.dat};

        \addplot[
          LightSlateGray!20,
          very thick,
          forget plot,
          name path=rew_sac
        ] table [x=samples, y=rew_sac] {exp/L1_cannon_F16.dat};

        \addplot[
          NavyBlue,
          very thick,
          name path=smoothed_spo
        ] table [x=samples, y=smoothed_spo] {exp/L1_cannon_F16.dat};

        \addplot[
          cyan,
          very thick,
          name path=smoothed_ppo
        ] table [x=samples, y=smoothed_ppo] {exp/L1_cannon_F16.dat};

        \addplot[
          LightSlateGray!80,
          very thick,
          name path=smoothed_sac
        ] table [x=samples, y=smoothed_sac] {exp/L1_cannon_F16.dat};
        
      \end{axis}
    \end{tikzpicture}
    \end{subfigure}
    \hspace{0.06cm}
    \begin{subfigure}{0.47\columnwidth}
    \centering
    \vspace{0.1cm}
    \begin{tikzpicture}
      \begin{axis}[
            title={A4 attack $\pi^{\mathrm{a}}_{a}$},
            title style={font=\bfseries\footnotesize, yshift=-0.75em},
            width=\columnwidth+0.8cm, height=3.5cm,
            xmin=-1, xmax=33000000,
            ymin=-6, ymax=8,
            xtick={0,10000000,20000000,30000000},
            xticklabels={$0\text{M}$, $10\text{M}$, $20\text{M}$, $30\text{M}$},
            scaled x ticks=false,
            ytick distance=4,
            enlargelimits=false,
            ticklabel style={font=\footnotesize},
            every axis plot/.append style={thick},
        ]

        \addplot[
          NavyBlue!20,
          very thick,
          forget plot,
          name path=rew_spo
        ] table [x=samples, y=rew_spo] {exp/L1_cannon_A4.dat};

        \addplot[
          cyan!20,
          very thick,
          forget plot,
          name path=rew_ppo
        ] table [x=samples, y=rew_ppo] {exp/L1_cannon_A4.dat};

        \addplot[
          LightSlateGray!20,
          very thick,
          forget plot,
          name path=rew_sac
        ] table [x=samples, y=rew_sac] {exp/L1_cannon_A4.dat};

        \addplot[
          NavyBlue,
          very thick,
          name path=smoothed_spo
        ] table [x=samples, y=smoothed_spo] {exp/L1_cannon_A4.dat};

        \addplot[
          cyan,
          very thick,
          name path=smoothed_ppo
        ] table [x=samples, y=smoothed_ppo] {exp/L1_cannon_A4.dat};

        \addplot[
          LightSlateGray!80,
          very thick,
          name path=smoothed_sac
        ] table [x=samples, y=smoothed_sac] {exp/L1_cannon_A4.dat};
        
      \end{axis}
    \end{tikzpicture}
    \end{subfigure}

    \begin{subfigure}{0.47\columnwidth}
    \centering
    \hspace*{-0.4cm}
    \vspace{0.1cm}
    \begin{tikzpicture}
      \begin{axis}[
            title={F16 defend $\pi^{\mathrm{f}}_{d}$},
            title style={font=\bfseries\footnotesize, yshift=-0.75em},
            width=\columnwidth+0.8cm, height=3.5cm,
            xmin=-1, xmax=33000000,
            ymin=-40, ymax=10,
            xtick={0,10000000,20000000,30000000},
            xticklabels={$0\text{M}$, $10\text{M}$, $20\text{M}$, $30\text{M}$},
            scaled x ticks=false,
            ytick distance=20,
            enlargelimits=false,
            ticklabel style={font=\footnotesize},
            every axis plot/.append style={thick},
        ]

        \addplot[
          NavyBlue!20,
          very thick,
          forget plot,
          name path=rew_spo
        ] table [x=samples, y=rew_spo] {exp/L1_escape_F16.dat};

        \addplot[
          cyan!20,
          very thick,
          forget plot,
          name path=rew_ppo
        ] table [x=samples, y=rew_ppo] {exp/L1_escape_F16.dat};

        \addplot[
          LightSlateGray!20,
          very thick,
          forget plot,
          name path=rew_sac
        ] table [x=samples, y=rew_sac] {exp/L1_escape_F16.dat};

        \addplot[
          NavyBlue,
          very thick,
          name path=smoothed_spo
        ] table [x=samples, y=smoothed_spo] {exp/L1_escape_F16.dat};

        \addplot[
          cyan,
          very thick,
          name path=smoothed_ppo
        ] table [x=samples, y=smoothed_ppo] {exp/L1_escape_F16.dat};

        \addplot[
          LightSlateGray!80,
          very thick,
          name path=smoothed_sac
        ] table [x=samples, y=smoothed_sac] {exp/L1_escape_F16.dat};
        
      \end{axis}
    \end{tikzpicture}
    \end{subfigure}
    \begin{subfigure}{0.47\columnwidth}
    \centering
    \vspace{0.1cm}
    \begin{tikzpicture}
      \begin{axis}[
            title={A4 defend $\pi^{\mathrm{a}}_{d}$},
            title style={font=\bfseries\footnotesize, yshift=-0.75em},
            width=\columnwidth+0.8cm, height=3.5cm,
            xmin=-1, xmax=33000000,
            ymin=-40, ymax=10,
            xtick={0,10000000,20000000,30000000},
            xticklabels={$0\text{M}$, $10\text{M}$, $20\text{M}$, $30\text{M}$},
            scaled x ticks=false,
            ytick distance=20,
            enlargelimits=false,
            ticklabel style={font=\footnotesize},
            every axis plot/.append style={thick},
        ]

        \addplot[
          NavyBlue!20,
          very thick,
          forget plot,
          name path=rew_spo
        ] table [x=samples, y=rew_spo] {exp/L1_escape_A4.dat};

        \addplot[
          cyan!20,
          very thick,
          forget plot,
          name path=rew_ppo
        ] table [x=samples, y=rew_ppo] {exp/L1_escape_A4.dat};

        \addplot[
          LightSlateGray!20,
          very thick,
          forget plot,
          name path=rew_sac
        ] table [x=samples, y=rew_sac] {exp/L1_escape_A4.dat};

        \addplot[
          NavyBlue,
          very thick,
          name path=smoothed_spo
        ] table [x=samples, y=smoothed_spo] {exp/L1_escape_A4.dat};

        \addplot[
          cyan,
          very thick,
          name path=smoothed_ppo
        ] table [x=samples, y=smoothed_ppo] {exp/L1_escape_A4.dat};

        \addplot[
          LightSlateGray!80,
          very thick,
          name path=smoothed_sac
        ] table [x=samples, y=smoothed_sac] {exp/L1_escape_A4.dat};
        
      \end{axis}
    \end{tikzpicture}
    \end{subfigure}
\caption{Training results of all $\pi_c$ on L1 across different tasks.}
\label{fig:low-level-training-results}
\end{figure}

To have a consistent comparison across algorithms, we consider training on the starting level L1. The results are plotted in Fig.~\ref{fig:low-level-training-results}. SPO and PPO achieved comparable performance, with SPO consistently outperforming PPO, while SAC performed worst. A reason for this is SAC's replay buffer that may store outdated transitions from obsolete opponent behaviors. Further, the shaping rewards~\eqref{eq:heading_rew_engage},~\eqref{eq:orientation_rew_engage}, and~\eqref{eq:pot_reward} may lead SPO and PPO to exploit these signals more directly, rather than SAC's exploration behavior, making the choice of \emph{on-policy} algorithms reasonable. Comparing F16 to A4, a slightly more stable learning curve for F16 agents is observed, likely due to their higher maneuverability. As agents should only destroy opponents when in attack mode, we consider $\pi_a$ of L4 in direct $1\text{-vs-}1$ dogfight against its version at L3 and observe the following results: $64\%$ win and $15\%$ loss for $\pi^{\text{f}}_{a}$, $56\%$ win and $19\%$ loss for $\pi^{\text{a}}_{a}$, with remaining outcomes being draws.

\subsection{High-Level Results}\label{sec:high-level-results}
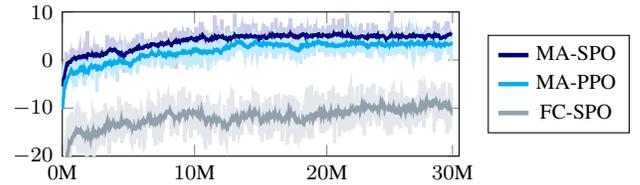
\begin{figure}[htb!]
    \centering
    \begin{tikzpicture}
      \begin{axis}[
            width=\columnwidth-2cm, height=3.5cm,
            xmin=-1, xmax=30000000,
            ymin=-20, ymax=10,
            xtick={0,10000000,20000000,29500000},
            xticklabels={$0\text{M}$, $10\text{M}$, $20\text{M}$, $30\text{M}$},
            scaled x ticks=false,
            ytick distance=10,
            enlargelimits=false,
            ticklabel style={font=\footnotesize},
            every axis plot/.append style={thick},
            legend style={
                at={(1.43,0.5)},
                anchor=east,
                draw=black, 
                fill=white,  
                font=\footnotesize,
            },
            legend columns=1,
            legend image post style={line width=1.7pt}, 
            legend image code/.code={
                \draw[#1] (0cm,0cm) -- (0.4cm,0.0cm); 
            }
        ]

        \addplot[
          LightSlateGray!20,
          very thick,
          forget plot,
          name path=rew_fc
        ] table [x=samples, y=rew_fc] {exp/hier.dat};

        \addplot[
          NavyBlue!20,
          very thick,
          forget plot,
          name path=rew_spo
        ] table [x=samples, y=rew_spo] {exp/hier.dat};

        \addplot[
          cyan!20,
          very thick,
          forget plot,
          name path=rew_ppo
        ] table [x=samples, y=rew_ppo] {exp/hier.dat};

        \addplot[
          NavyBlue,
          very thick,
          name path=smoothed_spo
        ] table [x=samples, y=smoothed_spo] {exp/hier.dat};
        \addlegendentry{MA-SPO};

        \addplot[
          cyan,
          very thick,
          name path=smoothed_ppo
        ] table [x=samples, y=smoothed_ppo] {exp/hier.dat};
        \addlegendentry{MA-PPO};

        \addplot[
          LightSlateGray!80,
          very thick,
          name path=smoothed_fc
        ] table [x=samples, y=smoothed_fc] {exp/hier.dat};
        \addlegendentry{FC-SPO};
        
      \end{axis}
    \end{tikzpicture}
    \caption{Training of $\pi_h$ through MA-SPO and MA-PPO and $\pi_{\text{\textit{\scriptsize fc}}}$ with FC-SPO against mixed strategy opponents (Sec.~\ref{sec:curriculum}).}
    \label{fig:hier_training}
\end{figure}

The high-level policies $\pi_{h}$ are trained under CTDE using the MA-SPO algorithm, which we compare against MA-PPO. SAC is excluded since it only supports continuous actions, making it incompatible with the option space $\mathcal{C}^i$ (Sec.~\ref{sec:high-level-policy}). To emphasize the strength of our hierarchical scheme, we omit an explicit ablation study of the components (CL, league-play) and instead directly compare it with a \emph{Fully Connected}~(FC) model $\pi_{\text{\textit{\scriptsize fc}}}$ under full observability (denoted FC-SPO), which inherently reflects their exclusion. While hierarchical policies $\pi^{\text{f}}_{h}$ and $\pi^{\text{a}}_{h}$ incorporate pre-trained maneuver policies $\pi_c$, $\pi_{\text{\textit{\scriptsize fc}}}$ is to engage directly against opponents with mixed strategies (Sec.~\ref{sec:curriculum}). The training results, shown in Fig.~\ref{fig:hier_training}, report the mean values for both aircraft types. MA-SPO performs slightly better than MA-PPO, validating the effectiveness of our adaptation of SPO to MA-SPO, while $\pi_{\text{\textit{\scriptsize fc}}}$ trained with FC-SPO fails to achieve positive combat results. This indicates that \emph{non}-hierarchical agents directly engaging strong opponents are likely to fail their mission, whereas our hierarchical agents effectively exploit strategic diversity through their maneuver options. While $\pi_{\text{\textit{\scriptsize fc}}}$ is \emph{decentralized}, a fully \emph{centralized} model is omitted, as it is less common in MARL and scales poorly. FC-PPO performed similar to FC-SPO and is also excluded.

The evaluation results in Table~\ref{tab:high-level-eval} cover various $n\text{-vs-}n$ scenarios, highlighting the flexibility of our approach via shared policies. We increase enemy aggressiveness by adjusting their mixed strategy to select $\pi_a (70 \text{\footnotesize{\%}}), \, \pi_e (20 \text{\footnotesize{\%}}), \, \pi_d (10 \text{\footnotesize{\%}})$. The horizon is set $10$ times higher as during training, as otherwise draws are mainly recorded in larger battle setups. Consistent with training results, we infer a clear superiority of MA-SPO and MA-PPO over FC-SPO. Our hierarchical MA-SPO agents show strong combat performance, achieving win rates above $80 \text{\footnotesize{\%}} $ even in competitive $10 \text{-vs-} 10$ scenarios. Inspecting low-level policy activations reveals the team-level strategy, which is approximated by averaging the chosen options of both aircraft types, while per-agent-type strategies can be directly extracted from the outputs of $\pi^{\text{f}}_{h}$ and $\pi^{\text{a}}_{h}$. As combat size increases, MA-SPO tends to become more cautious by favoring $\pi_e$ and $\pi_d$, whereas MA-PPO acts slightly more aggressively by selecting $\pi_a$ more often. A detailed analysis could examine under which conditions (e.g., distance $d$, aspect angle $\omega_a$, etc.) each option $\pi_c$ is chosen, revealing explanatory reasoning as similarly done in~\cite{combat_expl}, but this lies beyond the scope of this work.

\begin{table}[htb!]
\caption{Inference results and average option activation frequencies ($\bar{\pi}_{a/e/d}$) for different scenarios and models. All values are averaged over $1'000$ episodes and aircraft types.}
\begin{center}
\begin{tabular}{cccccccc}
    \Xhline{1.1pt}
    \noalign{\vskip 1.8pt}
    \textbf{Combat} & \textbf{Model} & Win & Draw & Loss & $\bar{\pi}_a / \pi_{\text{\textit{\scriptsize fc}}}$ & $\bar{\pi}_e$ & $\bar{\pi}_d$ \\ 
    \noalign{\vskip 1.8pt}
    \hline
    \noalign{\vskip 1.2pt}
    \multirow{3}{*}{$10\text{-vs-}10$} 
        & \textbf{MA-SPO} & $\mathbf{83\textbf{\scriptsize\%}}$ & $6\text{\scriptsize\%}$ & $11\text{\scriptsize\%}$ & $72\text{\scriptsize\%}$ & $21\text{\scriptsize\%}$ & $7\text{\scriptsize\%}$ \\
        & MA-PPO & $80\text{\scriptsize\%}$ & $7\text{\scriptsize\%}$ & $13\text{\scriptsize\%}$ & $61\text{\scriptsize\%}$ & $33\text{\scriptsize\%}$ & $6\text{\scriptsize\%}$ \\
        & FC-SPO & $0\text{\scriptsize\%}$ & $6\text{\scriptsize\%}$ & $94\text{\scriptsize\%}$ & $100\text{\scriptsize\%}$ & - & - \\
    \noalign{\vskip 1.2pt}
    \hline
    \noalign{\vskip 1.2pt}
        \multirow{3}{*}{$5\text{-vs-}5$} 
        & \textbf{MA-SPO} & $\mathbf{87\textbf{\scriptsize\%}}$ & $4\text{\scriptsize\%}$ & $9\text{\scriptsize\%}$ & $77\text{\scriptsize\%}$ & $18\text{\scriptsize\%}$ & $5\text{\scriptsize\%}$ \\
        & MA-PPO & $82\text{\scriptsize\%}$ & $6\text{\scriptsize\%}$ & $12\text{\scriptsize\%}$ & $59\text{\scriptsize\%}$ & $36\text{\scriptsize\%}$ & $5\text{\scriptsize\%}$ \\
        & FC-SPO & $0\text{\scriptsize\%}$ & $5\text{\scriptsize\%}$ & $95\text{\scriptsize\%}$ & $100\text{\scriptsize\%}$ & - & - \\
    \noalign{\vskip 1.2pt}
    \hline
    \noalign{\vskip 1.2pt}
        \multirow{3}{*}{$3\text{-vs-}3$} 
        & \textbf{MA-SPO} & $\mathbf{90\textbf{\scriptsize\%}}$ & $3\text{\scriptsize\%}$ & $7\text{\scriptsize\%}$ & $76\text{\scriptsize\%}$ & $20\text{\scriptsize\%}$ & $4\text{\scriptsize\%}$ \\
        & MA-PPO & $88\text{\scriptsize\%}$ & $3\text{\scriptsize\%}$ & $9\text{\scriptsize\%}$ & $56\text{\scriptsize\%}$ & $40\text{\scriptsize\%}$ & $4\text{\scriptsize\%}$ \\
        & FC-SPO & $0\text{\scriptsize\%}$ & $2\text{\scriptsize\%}$ & $98\text{\scriptsize\%}$ & $100\text{\scriptsize\%}$ & - & - \\
    \Xhline{1.1pt}
\end{tabular}
\label{tab:high-level-eval}
\end{center}
\end{table}

\section{Conclusion}\label{sec:conclusion}
This work introduced a HMARL framework for competitive multi-agent air combat scenarios, combining realistic flight physics, heterogeneous dynamics, a structured training pipeline with reward shaping, CL and league-play, and the adapted training algorithm MA-SPO. Our results show that hierarchical agents acquire coordinated strategies and exhibit strong resilience compared to non-hierarchical baselines. 

Nonetheless, some limitations remain. The reliance on fixed low-level controllers may restrict adaptability in highly dynamic or novel scenarios. While our setup focused on cannon-based close-range dogfights, further investigation is required to assess whether the hierarchical MA-SPO structure scales to more complex and long-range weapon systems. 

Future research will aim to strengthen tactical decision-making by incorporating advanced planning methods such as LightZero~\cite{lightzero}. In addition, our communication interface with VR-F enables direct interaction between human pilots and RL-controlled aircraft. By collecting explicit feedback from pilots, we plan to integrate imitation learning into the framework, fostering mixed human–RL teams and enabling new strategic maneuvers in operational training. Finally, our approach aims to ensure human oversight and to support trust and responsible deployment of intelligent agents in safety-critical domains.

\bibliographystyle{IEEEtran}
\bibliography{biblio}
\end{document}